\documentclass{article} 
\usepackage{iclr2027_conference,times}

\usepackage{amsmath,amsfonts,bm}

\def\eqref#1{equation~\ref{#1}}

\def\1{\bm{1}}

\DeclareMathAlphabet{\mathsfit}{\encodingdefault}{\sfdefault}{m}{sl}
\SetMathAlphabet{\mathsfit}{bold}{\encodingdefault}{\sfdefault}{bx}{n}

\usepackage{hyperref}       
\usepackage{url}            
\usepackage{booktabs}       
\usepackage{amsfonts}       
\usepackage{nicefrac}       
\usepackage{microtype}      
\usepackage{xcolor}         
\usepackage{graphicx}
\usepackage{amsmath}
\usepackage{array}
\usepackage{subcaption}
\usepackage{wrapfig}
\iclrfinalcopy

\title{Act4R: Action-Conditioned 4D Reconstruction for Kinematics-Free Control}

\author{Zhaoliang Zhang, Zhimin Shao, Xijun Liu  \\
Department of Electrical and Computer Engineering\\
Johns Hopkins University\\
\texttt{\{zzhan288, zshao14, xliu253\}@jhu.edu} \\
\And
Chenhao Yu, Cheng Peng \\
Department of Data Science \\
University of Virginia \\
\texttt{\{zzb9yb, xuz7wn\}@virginia.edu} \\
}

\begin{document}

\maketitle

\begin{abstract}

  Precise control of movement across various robotic embodiments is a longstanding challenge. Modeling complex robotic structures requires significant time, dedicated sensors, and manual labeling. Recent advancements in 4D representation learning offer an interesting alternative: can we leverage 4D features lifted from 2D observations to densely map a robot's spatial movement based on the action signal? To this end, we propose Act4R, a framework that builds on 3D foundation model to predict a robot's 3D scene flow with any given action. Act4R introduces a Spatial-Action Encoder, which encodes the current 2D robot state and its action signals into embeddings that represent future robot state in 3D. Then, both the current and future spatial embeddings are decoded into point maps to estimate action-conditioned scene flow. To effectively learn the mapping between action and future state, Act4R introduces spatial-action fusion in its transformer-based encoder, and a spatial alignment loss that guides the encoder for future flow prediction. After training on some samples of a given embodiment, Act4R can perform inverse control by optimizing the signal to achieve the desired scene flow. 
  We evaluate Act4R on a public real-world dataset and synthetic data collected in simulator. Experiments on various robotic platforms show that Act4R outperforms the previous SoTA in scene flow prediction. Our method also performs better in inverse control, both on the action accuracy metric and in closed-loop control in simulation.
  
  
  
  

  

\end{abstract}





\section{Introduction}

  Robotic control requires a reliable mapping from raw action signals to spatial movements. While most research uses well-measured and reliable embodiment with precise kinematics, e.g., Franka~\citep{haddadin2022franka}, Universal Robot~\citep{ur5e}, etc., there exist many more robots with which this mapping is not reliably specified. This can be due to manufacturing variation, limited kinematic sensing, or a design not originally intended for precise control, and is often true for low-cost or custom-built robots. 
  Relying on bespoke modeling to map raw signals to spatial movements~\citep{wong25} makes it difficult to support an open-ended range of embodiments.
  Our goal is a common visual calibration framework that learns this mapping for each embodiment, enabling downstream tasks without a manually specified kinematic model.
  
  
  Prior efforts to control robots with unmeasured kinematics have explored marker-based visual servoing and specialized machine learning architectures. 
  Early visual servoing techniques utilized high-contrast physical markers or LED arrays to track deformation \citep{Lagneau20, AlBeladi22, Bern20}; however, these methods often fail under occlusion or when the robot's surface undergoes significant warping. 
  Subsequent machine learning approaches attempted to bypass these physical constraints by training end-to-end regressors to map pixels directly to actuator commands \citep{Chin2020, Toshimitsu22}. Recently, Neural Jacobian Field (NJF) leverages radiance field as a representation to recover a robot's Jacobian matrix and action-conditioned scene flow at a pixel-level density~\citep{Li2025}, demonstrating a promising direction to map complex kinematics at full resolution. 
  However, training an NJF model requires substantial effort and imposes practical limitations.

  Instead of learning 3D representations from scratch, integrating strong 3D priors can improve data and learning efficiency of action-conditioned scene flow estimation. 
  Recent 3D Vision Foundation Models (VFM)~\citep{Wang24, Leroy24, Wang25vggt}   \begin{wrapfigure}{r}{0.7\columnwidth}
    \centering
    \includegraphics[width=\linewidth]{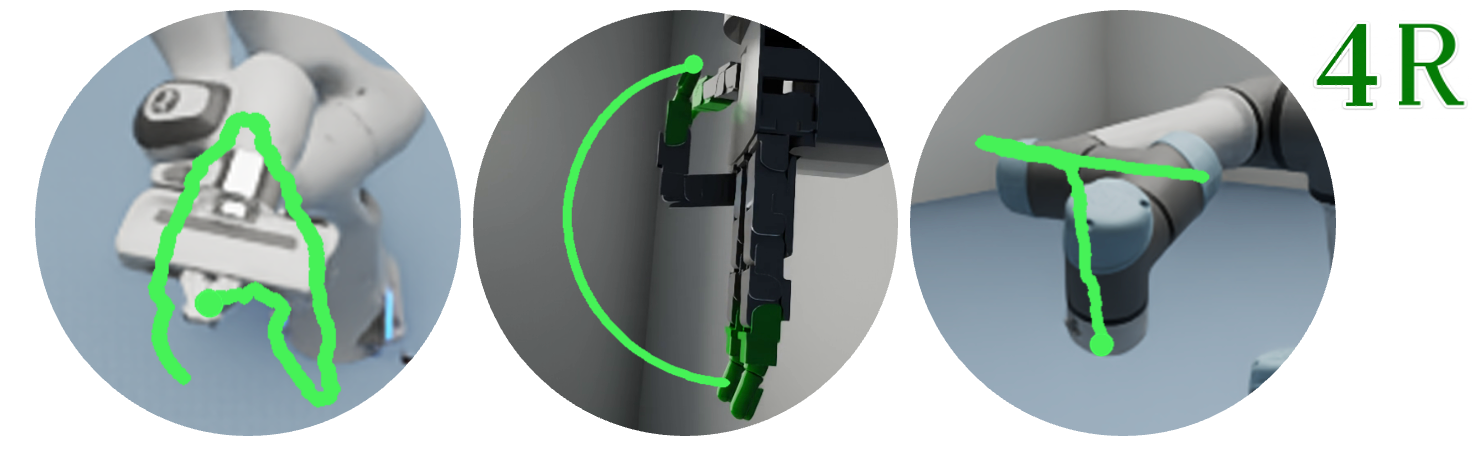}
    \caption{Robots can be controlled by Act4R without prior kinematics, as visualized by the green 2D tracks of three platforms.}
    \label{fig:intro_examples}
\end{wrapfigure}
  are trained on a large amount of 3D datasets, allowing them to internalize rich spatial representations. 
  These models exhibit remarkable robustness and zero-shot generalization capability across complex environments. 
  Building upon these foundations, recent works \citep{Zhang24, Feng25} have extended these capabilities to dynamic scenes, exploring the temporal evolution of geometry and motion. Particularly, St4RTrack~\citep{Feng25} demonstrates that scene flows can be estimated effectively by finetuning 3D VFMs on a small 4D labeled dataset. This raises an interesting question: can 3D VFMs be adapted to a novel modality, i.e., the robot's raw action signal, to estimate its future scene flow and enable action-to-motion calibration across different embodiments?

  To this end, we introduce \textbf{Act4R}, a method for action-to-motion calibration that predicts a robot's future \textit{spatial} movement from the current 2D observation and an input action command. Rather than requiring a manually specified kinematic model, Act4R learns this mapping from observations and action signals for each embodiment. The main challenge for Act4R lies in finding an \textit{efficient} and \textit{performant} mapping from action signals to spatial representations. Previous works in action encoding, e.g., for Vision-Language-Action (VLA) models, focus on descriptive, language-based actions~\citep{kim2025openvla,Qu25, Li2025SF}. Gr00t N1~\citep{bjorck2025gr00t} introduces a three-layer MLP to convert action into the token space; however, these designs are ineffective at inferring spatial concepts from raw action signals. In contrast, 3D VFMs encode images into spatial embeddings, but have not been applied to action signals. Act4R introduces a novel Spatial-Action Encoder (SAE) module, which cross-attends the spatial embedding from a 2D observation and an action signal. SAE outputs an updated spatial embedding, which is then fed to a 3D VFM decoder to generate point maps for scene flow estimation. To ensure that SAE is learning a useful representation for the 3D VFM decoder, Act4R imposes a spatial alignment loss on SAE. Specifically, SAE's output should align with the spatial embedding of the \textit{future} image. 

  During training, Act4R internalizes the robot's geometric deformations by observing a robot execute a diverse range of actions from various angles. For geometric supervision, we use sensor measured depth. A further ablation study also shows that depth estimated by a 3D VFM can also be used for effective control mapping, as demonstrated by the green tracks produced by different robots in Fig.~\ref{fig:intro_examples}.
  Once the VFM is finetuned for a specific robot, Act4R can be used for inverse control. 
  On data collected from four real-world robotic platforms, we demonstrate that Act4R improves scene flow prediction and inverse action estimation while requiring less training time than NJF~\citep{Li2025}. 
  To evaluate the same calibration framework across additional embodiments, we collect a simulation dataset covering three robotic embodiments and establish a common evaluation protocol for unknown-embodiment control. Experiments on these embodiments demonstrate effective closed-loop inverse control, complementing the offline evaluation on real-world data. 

  In summary, our main contributions are:
  \begin{enumerate}
    \item We propose \textbf{Act4R}, an action-conditioned scene flow estimation framework for calibrating action-to-motion mappings across robotic embodiments without manually specified kinematic models. 
    \item Within Act4R, we propose a \textbf{Spatial-Action Encoder} (SAE). SAE introduces a spatial-action-fusion mechanism to ensure spatial and action embeddings are modeled jointly; a spatial alignment loss is introduced to ensure a correct mapping from action signals to spatial embeddings is learned.
    \item Based on experiments over four real-world robotic platforms, our framework significantly improves upon previous approaches~\citep{Li2025}.
    We additionally collect a simulation dataset and establish a shared evaluation protocol across three robotic embodiments, providing a portable benchmark for unknown-embodiment control.
    
  \end{enumerate}

\section{Related Works}
\label{sec:relatedwork}

\subsection{Robotic Control Mapping}
 In conventional robotic manipulation, the control framework typically relies on a manually designed Jacobian system that maps control commands to the corresponding velocities of the end-effector \citep{whitney1969resolved}. 
 With the rapid development of deep learning, recent works, as surveyed in \citep{weinberg2024survey} have explored modeling robotic systems and their Jacobians with neural networks. Most work require a bespoke modeling scheme for specific robotic platforms, e.g., exhaustive joint-level $\text{DoF}$ annotations to map actuator space to Cartesian space. 
 To bypass these modeling hurdles, some frameworks \citep{Yip14, Lagneau20, Bern20, AlBeladi22, Toshimitsu22} utilize marker-based visual servoing to derive low-level control signals. 
 While effective for tracking, these methods rely on manual instrumentation, such as ArUco markers or color-coded segments, which limits their applicability to unstructured real-world scenarios where such modifications are impractical or impossible. 
 
 In recent years, 
 Vision-Language-Action (VLA) models have been introduced to ground high-level semantic knowledge into low-level robotic control. 
 While early VLA models relied primarily on 2D semantic alignments, recent advancements have introduced more sophisticated spatial understanding capabilities~\citep{Qu25,kim2025openvla, Li2025SF}. \cite{huang2026pointworld} predict action-conditioned 3D scene flow using robot point-flow actions constructed from known robot geometry and forward kinematics. We note that these works are built upon rigid-body robotic platforms with well-articulated kinematics and controls and cannot be transferred to platforms where control is unknown. 
 
 
 Parallel to high-level VLA reasoning, Neural Jacobian Field (NJF) \citep{Li2025} has emerged as a novel paradigm for inverse robotic control by representing kinematics as a continuous, coordinate-based Neural Field $\mathbf{J}(\mathbf{x}) = \partial \mathbf{x} / \partial \mathbf{u}$. NJF is optimized by observing robotic movement across various view angles and updates itself through ray-tracing an action-conditioned, differentiable displacement field~\citep{Pumarola21,Park21}. By modeling every voxel of the robot, NJF provides a generalizable approach for high DoF inverse control. 
 Act4R builds upon this paradigm but in a different way. Act4R leverages 3D VFMs to represent spatial relationship between an image and a control signal; as such the choice of representation is in point maps and not Neural Fields. 
 \vspace{-5pt}
\subsection{Progress in 3D Vision Foundation Models}
 Significant progress has been made recently in feed-forward neural networks that can estimate geometry, camera poses and feature tracks from two-view or multi-view inputs. DUSt3R and MASt3R~\citep{Wang24,Leroy24} introduce the point-map representation and canonical camera coordinate framework that enables efficient learning of geometry. Subsequent work extends this formulation to continuous captures~\citep{wang2025continuous}, multi-view~\citep{Wang25vggt,wang2026pi,lin2025depth}, over long sequences~\citep{zhuo2025streaming,deng2025vggt}, etc.
 Moving beyond 3D modeling, Shape of Motion \citep{Wang25} and Dynamic Point Maps \citep{Sucar25} emphasize the coupling of geometry and motion, providing versatile representations for reconstructing complex deformations. 
 More recently, foundation models like MonST3R \citep{Zhang24} and St4RTrack \citep{Feng25} have simplified the estimation of geometry in the presence of motion. 
 By leveraging large-scale visual priors, these frameworks enable simultaneous 4D reconstruction and point-to-point tracking, providing the dense supervision required for the training of our action-conditioned model. Act4R is built upon the general encoder-decoder Transformer architecture of these 3D/4D and inherits some of the prior weights; however, Act4R takes an image and an action as inputs, instead of an image pair. As such, Act4R introduces its own encoder to convert action information into spatial embeddings. 
 \vspace{-5pt}

\section{Methodology}
\label{sec:method}

  \begin{figure*}[t]
      \centering
      \includegraphics[width=\textwidth]{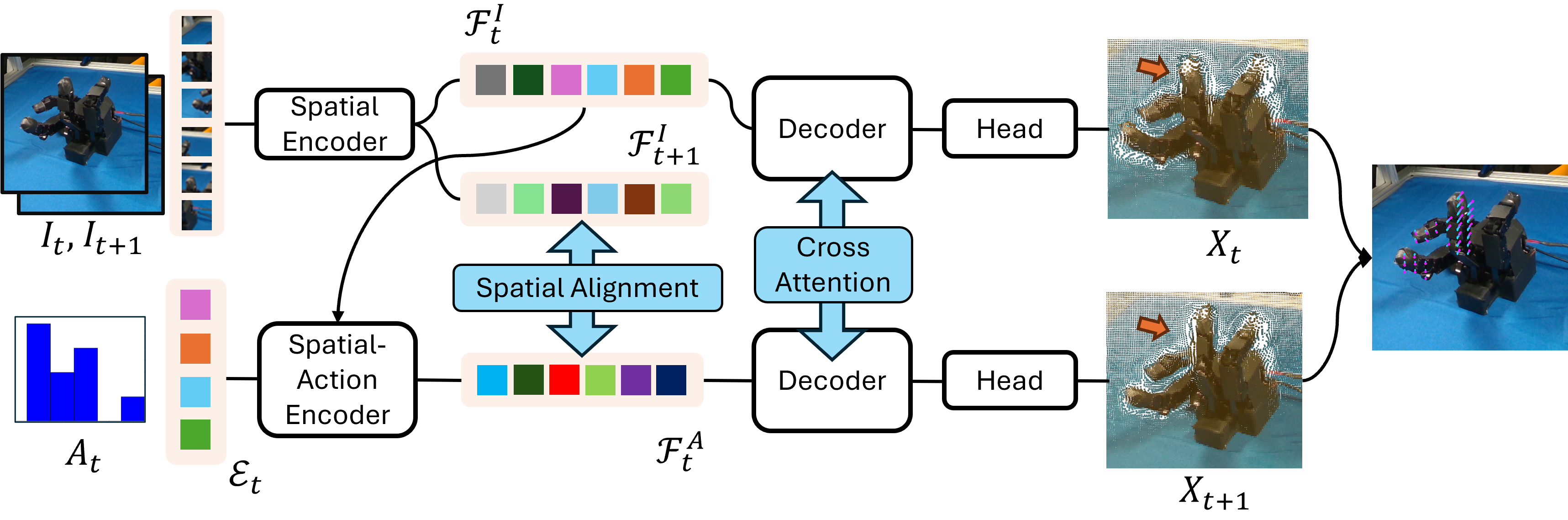}
      \caption{\textbf{Method Overview}: Given a 2D observation of the robot $I_t$ and the action $A_t$, Act4R introduces a Spatial-Action Encoder to estimate the future placement of the robot at time $t+1$. A learnable action embedding $\mathcal{E}_t$ is introduced to improve efficiency. \textbf{At training time}, SAE's output feature $\mathcal{F}^A_t$ is aligned with the future spatial feature $\mathcal{F}^I_{t+1}$. At inference time, $I_{t+1}$ is not used.}
      \label{fig:pipeline}
  \end{figure*}

  Act4R seeks to map actions to a robot's spatial movement from observations only. As shown in Fig. ~\ref{fig:pipeline}, Act4R predicts future robotic state by taking the current image $I_t$ and a control action $A_t$ as inputs. We detail the overall problem setting and Act4R framework in Sec. \ref{sec:ps}.
  Within Act4R, we introduce a Spatial-Action Encoder (SAE), a specialized action encoder to map control commands into high-dimensional spatial embeddings, as detailed in Sec. \ref{sec:sae}.
  At training time, we use a spatial alignment loss to constrain the output from SAE with the spatial embeddings of the future image $I_{t+1}$, described in Sec. \ref{sec:fssa}. 
  Following this alignment, the decoder is trained end-to-end under dense reconstruction and tracking supervision. 
  Finally, we invert the model to derive the optimal action required to achieve the desired robotic motion. 

\subsection{Problem Settings and the Act4R Framework}
\label{sec:ps}
  At time $t$, we define the 3D pointmap $\mathbf{X}_t$ of the robot as its system state, and the action vector $A_t$ as the control command. In reality, it is often hard to have direct access to $\mathbf{X}_t$; instead, we can easily have access to a 2D observation $I_t$ of the robot. The goal of Act4R is to predict the current and future state of the robot $\mathbf{X}_{t, t+1}$ by training on a series of $I_t$, $A_t$, and $I_{t+1}$ collected over time. Subsequently, we can obtain the scene flow $\mathcal{S}$ and manipulate the robot based on predicted scene flow.


  Our approach leverages 3D VFMs priors~\citep{Feng25, Wang24, Wang25vggt},
  which train a neural network $f_\theta$ to regress dense 3D point maps from an image pair $(\mathbf{I}_i, \mathbf{I}_j)$:
  \begin{equation}
      \mathit{f}_{\theta}(\mathbf{I}_i, \mathbf{I}_j) = \{\mathbf{X}^{i}_{i}, \mathbf{X}^{i}_{j}\},
  \end{equation}
  $\mathbf{X}^i_j \in \mathbb{R}^{N \times 3}$ represents a dense set of 3D points from $I_j$ in the camera coordinate frame of $I_i$. Generally speaking, 3D VFMs can be split into a spatial encoder $f_{\theta_{enc}}$ and decoder $f_{\theta_{dec}}$, i.e.:
    \begin{equation}
      \mathit{f}_{\theta_{enc}}(\mathbf{I}_i) = \mathcal{F}^{I}_{i}, \mathit{f}_{\theta_{enc}}(\mathbf{I}_j) = \mathcal{F}^{I}_{j}, \mathit{f}_{\theta_{dec}}(\mathcal{F}^{I}_{i},\mathcal{F}^{I}_{j}) = \{\mathbf{X}^{i}_{i}, \mathbf{X}^{i}_{j}\},
  \end{equation}
  where $\mathit{f}_{\theta_{enc}}$ first lifts images into spatial embeddings, and $\mathit{f}_{\theta_{dec}}$ performs cross attention between two embeddings to generate 3D point maps.  
  Exploiting the flexibility of Transformer-based architectures, which process inputs as sequences of tokens, we substitute the target image $\mathbf{I}_{j}$ with an action vector $\mathbf{A}_t$. 
  This modification enables the framework to generate a 3D point map for the current observation $\mathbf{I}_t$ conditioned on $\mathbf{A}_t$, the signal that implies future robotic state.
  Because the camera pose remains static between observations, the coordinate frame is static, and the predicted point map varies only in the temporal dimension. 
  Thus, our model formulation becomes:
  \begin{equation}
      \mathit{f}_{\theta_{enc}}(\mathbf{I}_t) = \mathcal{F}^{I}_{t}, \mathit{f}^{\textrm{SAE}}_{\theta_{enc}}(\mathbf{A}_t, \mathcal{F}^{I}_{t}) = \mathcal{F}^{A}_{t}, \mathit{f}_{\theta_{dec}}(\mathcal{F}^{I}_{t},\mathcal{F}^{A}_{t}) = \{\mathbf{X}_{t}, \mathbf{X}_{t+1}\}.
  \end{equation}
   As shown in Fig.~\ref{fig:pipeline}, Act4R uses an SAE to lift $A_t$ into spatial embedding. It is then fed into a 3D VFM decoder with the spatial embedding from $I_t$, and produces $X_{t,t+1}$.

\subsection{Spatial-Action Encoder}
\label{sec:sae}
How can we find the proper mapping from action signals $A_t$ to the future 3D robotic state $\textbf{X}_{t+1}$, given the current state in $I_t$ and $\mathcal{F}^{I}_{t}$? A naive approach may be to encode action signals into action embeddings, and train a decoder architecture that cross-attends with $\mathcal{F}^{I}_{t}$. While this may work well if there exists a large amount of image-action pairs and their 3D ground truth, we generally do not have enough data to train a decoder from scratch to interpret the spatial-action relationship. As such, we leverage existing spatial prior in 3D VFMs, where the decoder are trained to take spatial embeddings encoded from two images. 

\begin{figure*}[t]
      \centering
      \includegraphics[width=\textwidth]{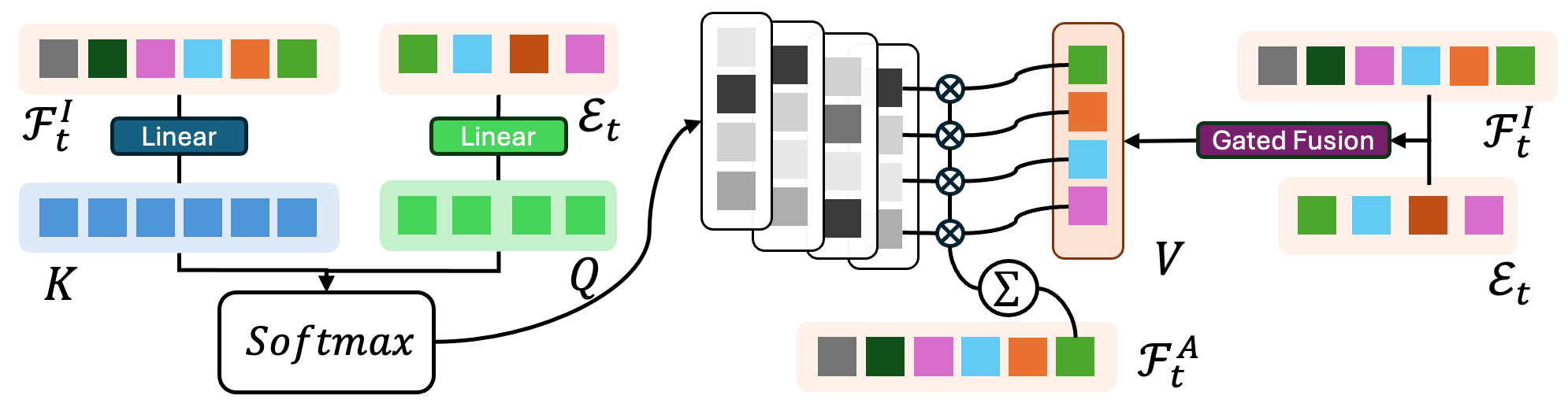}
      \caption{The architecture of Spatial-Action Encoder.}
      \label{fig:sae}
      \vspace{-10pt}
  \end{figure*}

As shown in Fig.~\ref{fig:sae}, we move the fusion of $A_t\in\mathbb{R}^M$ and $\mathcal{F}^{I}_{t}\in\mathbb{R}^{N\times C}$ into a Spatial-Action Encoder. SAE first passes $A_t$ through an MLP to construct the action embedding $\mathcal{E}_t \in \mathbb{R}^{M\times C}$. Similar to conventional Transformer architecture, $\mathcal{F}^{I}_{t}$ and $\mathcal{E}_t$ are linearly projected to form the Key and Query tokens. Attention weights $W\in\mathbb{R}^{M\times N}$ are computed through a dot product and softmax operation. Conventional Cross Attention mechanism typically expresses Value token in one modality, e.g. image tokens query on text token space. In our case, both action $\mathcal{E}_t$ and $\mathcal{F}^{I}_{t}$ contain important spatial information about $\textbf{X}_{t+1}$; therefore, we construct the Value token $V\in\mathbb{R}^{N\times C}$ by a gated channel-wise fusion of the image tokens and a motion-informed context. We average the action tokens $\mathcal{E}_{t}\in\mathbb{R}^{M\times C}$ over the $M$ joints, broadcast the result to the $N$ image patches, and apply a linear projection $W_{v}\in\mathbb{R}^{C\times C}$ to obtain $\tilde{\mathcal{E}}_{t}\in\mathbb{R}^{N\times C}$. A per-patch gate $g=\sigma(\mathrm{MLP}([\mathcal{F}^{I}_{t};\tilde{\mathcal{E}}_{t}]))\in\mathbb{R}^{N\times 1}$ then interpolates between the two sources as $V=(1-g)\odot\mathcal{F}^{I}_{t}+g\odot\tilde{\mathcal{E}}_{t}$. After multiplying the attention weights with the Value tokens, we obtain an encoded action token $\mathcal{E}_{t'}\in\mathbb{R}^{M\times C}$. We then feed $\mathcal{E}_{t'}$ as Key and $\mathcal{F}^{I}_{t}$ as Query in the next layer. In the second layer the Key already has length $M$, so $V\in\mathbb{R}^{M\times C}$ is formed with the same gated fusion, now interpolating $\mathcal{E}_{t'}$ and a per-joint projection of $\mathcal{E}_{t}$; we do not average over joints or concatenate $\mathcal{F}^{I}_{t}$. This fusion mechanism allows action embedding to be continuously integrated into the spatial embedding, rather than aggregating only on Value tokens from $\mathcal{F}^{I}_{t}$. The final output is denoted as $\mathcal{F}^{A}_{t}\in\mathbb{R}^{N\times C}$.

\subsection{Future State Spatial Alignment}
\label{sec:fssa}
   
A direct "cold start" through end-to-end training with $\mathbf{A}_t$ and SAE risks catastrophic interference with the pre-trained weights, potentially distorting the high-fidelity spatial representations internalized during large-scale pre-training. 
To preserve these geometric priors, we introduce a latent alignment constraint. At \textit{training time}, we introduce an additional spatial embedding $\mathcal{F}^{I}_{t+1}$, which represents the pseudo ground truth spatial embedding that we hope SAE can learn from. 

To this end, we first construct an oracle version of 3D VFMs $\mathit{f}_{\phi}$ by finetuning on the following predictions:
  \begin{equation}
      \mathit{f}_{\phi}(\mathbf{I}_t, \mathbf{I}_{t+1}) = \{\mathbf{X}_{t}, \mathbf{X}_{t+1}\}.
  \end{equation}
Act4R initializes its spatial encoder and decoder weights from $\mathit{f}_{\phi}$, and freezes its spatial encoder during training. We extract the intermediate spatial embedding $\mathcal{F}^{I}_{t+1}$ by passing $I_{t+1}$ to the spatial encoder, and force $\mathcal{F}^{A}_{t}$ to match $\mathcal{F}^{I}_{t+1}$. By doing so, the action encoder inherits the foundation model's visually grounded spatial logic. 
This ensures that control inputs are mapped into a physically meaningful 3D manifold, maintaining the integrity of the model's predictive capacity. The alignment loss is formulated as:
\begin{equation}
      \mathcal{L}_{feat} = \left\| \mathcal{F}^{A}_{t} - \mathcal{F}^{I}_{t+1} \right\|_2^2.
  \end{equation}

\subsection{Action-Conditioned 4D Prediction} 

  \textbf{Scene Flow Loss}.
  To supervise the geometric deformation induced by the action, we utilize the state-of-the-art point tracking framework CoTracker \citep{karaev23cotracker, Karaev25} as pseudo ground truth. 
  Because the model predicts 3D coordinates while CoTracker provides 2D temporal correspondences ($x_{t,n}^{gt}, x_{t+1,n}^{gt}$), we lift both endpoints with the depth maps $X^{\mathrm{gt}}_{t,n} = \Pi^{-1}(x_{t,n}^{gt}; D_{t})$, $\mathcal{S}_{gt,n} = X^{\mathrm{gt}}_{t+1,n} - X^{\mathrm{gt}}_{t,n}$, where $\Pi^{-1}(\cdot; D)$ denotes the back-projection with depth $D$. 
  The predicted scene flow at CoTracker correspondence $n$ is $\mathcal{S}_n = X_{t+1,n} - X_{t,n}$. Since the foundation model is scale-ambiguous, we first align the scale of gt and predicted  over all pixels $\mathcal{V}$ with valid depth $\hat{\mathcal{S}_n}$ and calculate loss:  
  \begin{equation}
  \label{equ:flowalign}
      \hat{\mathcal{S}_n} = s_{flow}\mathcal{S}_n, \qquad s_{flow} = \frac{\sum_{p \in \mathcal{V}} X_{t,p}^{\top} X^{\mathrm{gt}}_{t,p}}{\sum_{p \in \mathcal{V}} \left\| X_{t,p} \right\|_{2}^{2}}, \qquad \mathcal{L}_{flow} = \frac{1}{N} \sum_{n=1}^{N} \left| \mathcal{S}_{gt,n} - \hat{\mathcal{S}_n} \right|_2.
  \end{equation}
  \vspace{-10pt}

  \textbf{Trajectory Loss}. 
  We also use 2D track generated by CoTracker as a supervision directly. 
  The post-action point map is projected onto image plane, $x_{t+1} = \Pi_{t+1}\left(\mathbf{X}_{t+1} \right)$.
  The same scale ambiguity remains in the image plane, so we rescale the predicted track about the image center $c$ and then calculate the trajectory loss $\mathcal{L}_{traj}$:
  \begin{equation}
    \hat{x}_{t+1} = (x_{t+1} - c)\, s_{traj} + c, s_{traj} = \frac{1}{N} \sum_{n=1}^{N} \frac{\left\| x_{t+1,n}^{gt} - c \right\|_{2}}{\left\| x_{t+1,n} - c \right\|_{2}},  \mathcal{L}_{traj} = \frac{1}{N}\sum_{n=1}^{N} \left\|x_{t+1,n}^{gt} - \hat{x}_{t+1,n}\right\|^2.
  \end{equation}
Additionally, we follow \cite{Feng25} in applying a depth loss $\mathcal{L}_{depth}$ to supervise 3D geometry, and a consistency loss $\mathcal{L}_{cons}$ to promote temporal stability. More details are included in the Appendix. 
  The overall loss during training is:
  \begin{equation}
  \label{totaloss}
      \mathcal{L} = \lambda_{feat}\mathcal{L}_{feat} + \lambda_{flow}\mathcal{L}_{flow} + \lambda_{traj}\mathcal{L}_{traj} + \lambda_{depth}\mathcal{L}_{depth} + \lambda_{cons}\mathcal{L}_{cons}
  \end{equation}

\subsection{Inverse Control from Future Scene Flow}
\label{sec:modelapprox}
Once Act4R is trained to model the future scene flow, we can use it for inverse control. 
Unlike conventional kinematics methods, Act4R models a robot's movement at every point and applies to a wide range of robots without further efforts.
Consider that we want the robot to move from a current state to a desired state.
This can be done by establishing the desired 2D or 3D flow
$\mathcal{S}_{tgt}$ from the current observation $I_t$. We use $\ell{2}$ optimization objective that finds the best $A^*$ to achieve $\mathcal{S}_{tgt}$ through $\mathit{f}_{\theta}$. In 3D form, this is:

  \vspace{-18pt}
  \begin{equation}
  \label{eq:inverse_3D}
      A^*_t = \operatorname*{argmin}_{A_t} \sum_{n=1}^{N} \left\| \Delta^{3D}\!\left(\mathit{f}_{\theta}(I_t, A_t)\right) - \mathcal{S}^{3D}_{tgt,n} \right\|_2^2,
    \Delta^{3D}(X_t, X_{t+1}) = X_{t+1,n} - X_{t,n}.
  \end{equation}
  \vspace{-5pt}
  We can also project the 3D flow from Act4R to fit the 2D tracks.
  \vspace{-5pt}
  \begin{equation}
  \label{eq:inverse_2D}
      A^*_t = \operatorname*{argmin}_{A_t} \sum_{n=1}^{N} \left\| \Delta^{2D}\!\left(\mathit{f}_{\theta}(I_t, A_t)\right) - \mathcal{S}^{2D}_{tgt,n} \right\|_2^2,
    \Delta^{2D}(X_t, X_{t+1}) = \Pi(X_{t+1,n}) - \Pi(X_{t,n})
  \end{equation}  
 We can implement this by freezing the network parameters $\mathit{f}_{\theta}$ and use popular optimizers, e.g., Adam~\citep{kingma2014adam} to minimize Eq. \ref{eq:inverse_3D} or Eq. \ref{eq:inverse_2D}. During inverse control, only the current observation and the desired motion flow are required.

\section{Experiments}
\label{sec:exp}


\subsection{Experiment Setup}

\begin{table}[!htb]
  \centering
  \small
  \begin{tabular}{l ccccc}
    \toprule
    & \multicolumn{5}{c}{3D Flow Error (mm) $\downarrow$} \\
    \cmidrule{2-6}
    Robot & Act4R & $\text{Act4R}_{\text{NSA}}$ & $\text{Act4R}_{\text{N3D}}$ & $\text{Act4R}_{\text{Gr00t}}$ & NJF~\citep{Li2025} \\
    \midrule
    Allegro        & \textbf{6.715} & 7.947 & 112.113 & 10.704 & 11.607 \\
    HSA            & \textbf{5.136} & 5.740 & 58.704 & 5.757 & 7.420 \\
    Toy Arm        & \textbf{6.820} & 8.199 & 43.755 & 8.153 & 9.398 \\
    Pneumatic      & \textbf{4.603} & 5.130 & 43.955 & 5.257 & 10.744 \\
    Allegro Sim    & \textbf{3.765} & 4.283 & 74.590 & 8.333 & 17.666 \\
    Franka Sim        & \textbf{6.675} & 6.766 & 9.988 & 14.571 & 35.155 \\
    UR5e Sim          & \textbf{4.787} & 11.295 & 146.750 & 15.040 & 32.246 \\
    \midrule
    Mean           & \textbf{5.500} & 7.051 & 69.979 & 9.688 & 17.748 \\
    \bottomrule
  \end{tabular}
  \caption{Performance comparison of 3D flow error.}
  \label{tab:flow3d}
  \vspace{-10pt}
\end{table}

  \textbf{Real World Dataset.} We use the dataset released by \cite{Li2025}, which provides a collection of robotic platforms and sensor data. 
  The dataset features four distinct robot types to evaluate generalization across varied embodiments: 1. Allegro Hand \citep{wonikhand}, a commercial four-fingered robotic hand, 2. HSA \citep{Lipton18}, a unique platform utilizing four deformable, helical-actuator-driven legs, 3. Pneumatic Hand \citep{Matusik24}, a 3D-printed pneumatic hand combining structural rigidity with soft-robotic flexibility, and 4. a low-cost, multi-jointed robotic arm designed for educational use \citep{Lapeyre15}.

  \textbf{Simulation Dataset.}
  Following the same settings, we collect three additional embodiments in simulators. In IsaacSim simulator \citep{gao2026nvidia}, we use 1. Franka~\citep{haddadin2022franka}, 2. Allegro Hand~\citep{wonikhand} and 3. UR5e~\citep{ur5e} robots as the robotic platforms. 
  
  All real and simulated datasets include multi-view RGB video frames and depth maps, along with the control signal for each robot. The depth maps are measured by dedicated sensors for real data, and environment ground truth for synthetic data.
  For each robotic platform, the dataset comprises several hundred individual action sequences, every action sequence consists of 8-10 frames. 
  We use 90\% of the sequences as training and validation set and the rest as test set. 

  
  \textbf{Training and Inference.} All experiments are conducted with four NVIDIA RTX A40 GPUs. 
  We leverage St4RTrack~\citep{Feng25}, a recent work on 4D reconstruction based on 3D VFMs~\citep{Wang24,Leroy24}, as our base model. 
  The Act4R architecture is trained for 10 epochs. 
  We train our model individually for every robotic platform, taking approximately 16 hours. For baselines, NJF \citep{Li2025} is trained for 32 hours.
  


  

\newcommand{\rowlabel}[1]{%
    \rotatebox{90}{%
        \begin{minipage}{2cm} 
            \centering\textbf{#1}
        \end{minipage}%
    }%
}

\subsection{Flow Prediction Accuracy}

\begin{figure}[htbp]
    \centering
    \setlength{\tabcolsep}{2pt} 
    \renewcommand{\arraystretch}{0.9} 
    \begin{tabular}{>{\raggedright\arraybackslash}p{0.9cm} cccc}
        & \small \textbf{Allegro} & \small \textbf{HSA} & \small \textbf{ToyArm} & \small \textbf{Pneumatic} \\[0.5ex]
        
        \rowlabel{GT} &
        \begin{subfigure}{0.22\textwidth}
            \centering
            \includegraphics[width=\linewidth]{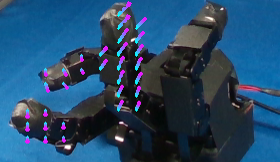}
        \end{subfigure} &
        \begin{subfigure}{0.22\textwidth}
            \centering
            \includegraphics[width=\linewidth]{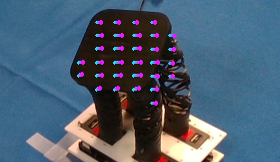}
        \end{subfigure} &
        \begin{subfigure}{0.22\textwidth}
            \centering
            \includegraphics[width=\linewidth]{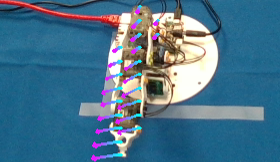}
        \end{subfigure} &
        \begin{subfigure}{0.22\textwidth}
            \centering
            \includegraphics[width=\linewidth]{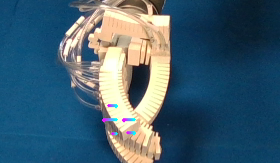}
        \end{subfigure} \\[0ex]

        \rowlabel{NJF~\citep{Li2025}} &
        \begin{subfigure}{0.22\textwidth}
            \centering
            \includegraphics[width=\linewidth]{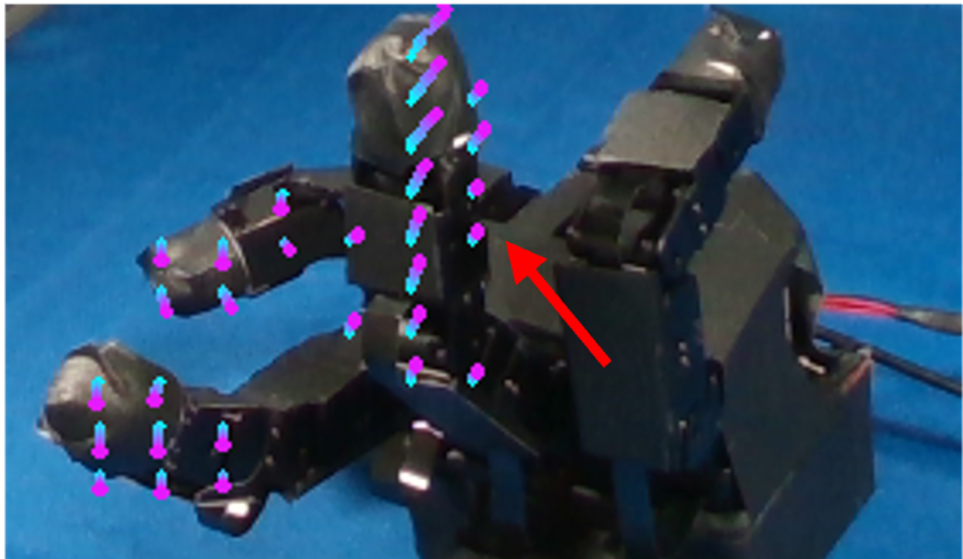}\\[-2pt]
            \scriptsize Mean Pixel Error: 3.619
        \end{subfigure} &
        \begin{subfigure}{0.22\textwidth}
            \centering
            \includegraphics[width=\linewidth]{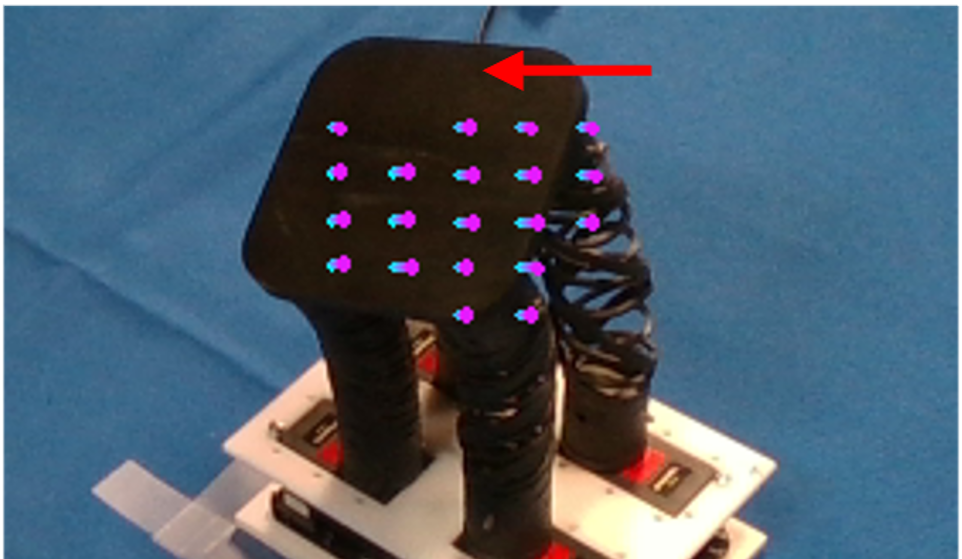}\\[-2pt]
            \scriptsize 2.539
        \end{subfigure} &
        \begin{subfigure}{0.22\textwidth}
            \centering
            \includegraphics[width=\linewidth]{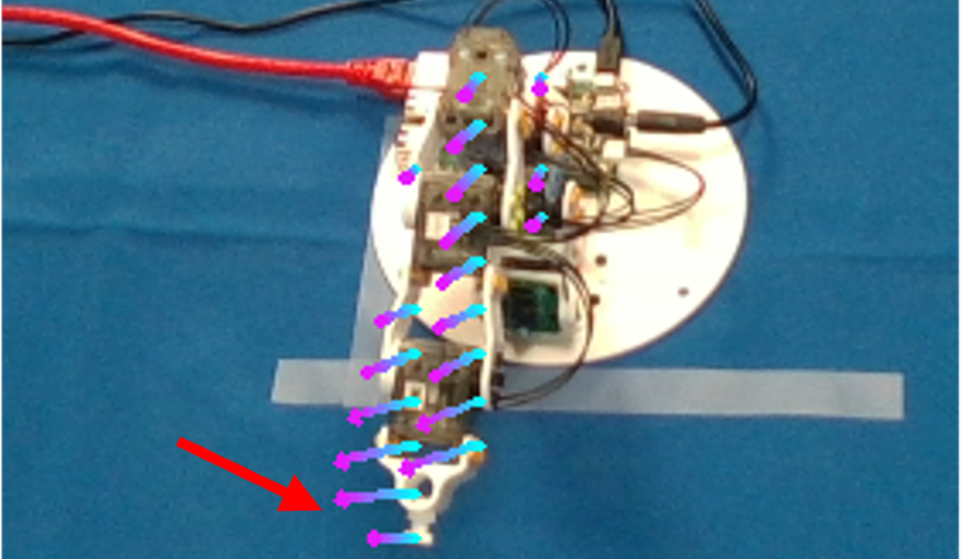}\\[-2pt]
            \scriptsize 5.794
        \end{subfigure} &
        \begin{subfigure}{0.22\textwidth}
            \centering
            \includegraphics[width=\linewidth]{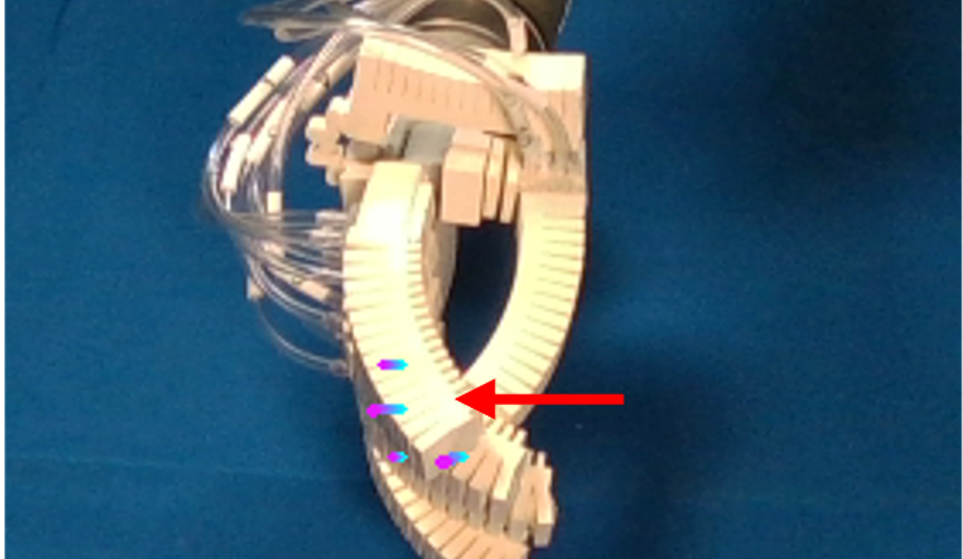}\\[-2pt]
            \scriptsize 2.157
        \end{subfigure} \\[0ex]

        \rowlabel{Act4R} &
        \begin{subfigure}{0.22\textwidth}
            \centering
            \includegraphics[width=\linewidth]{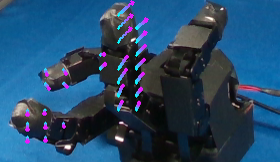}\\[-2pt]
            \scriptsize Mean Pixel Error: \textbf{2.706}
        \end{subfigure} &
        \begin{subfigure}{0.22\textwidth}
            \centering
            \includegraphics[width=\linewidth]{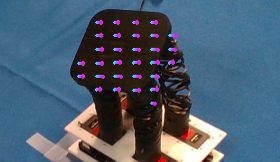}\\[-2pt]
            \scriptsize \textbf{1.101}
        \end{subfigure} &
        \begin{subfigure}{0.22\textwidth}
            \centering
            \includegraphics[width=\linewidth]{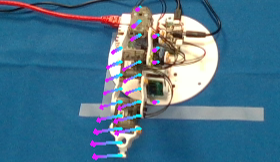}\\[-2pt]
            \scriptsize \textbf{3.565}
        \end{subfigure} &
        \begin{subfigure}{0.22\textwidth}
            \centering
            \includegraphics[width=\linewidth]{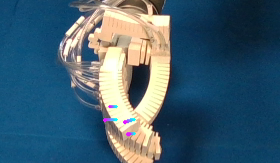}\\[-2pt]
            \scriptsize \textbf{1.473}
        \end{subfigure} \\
    \end{tabular}
    \caption{Predicted tracks across different experimental setups. GT denotes ground truth. Compared to \cite{Li2025}, Act4R is much more accurate at predicting action-conditioned flow.}
    \label{fig:track}
\end{figure}
\vspace{-10pt}
  For scene flow evaluation, we follow a bidirectional protocol, evaluating every image pair in both forward and backward directions. 
  During testing, the model takes $I_t$ and $A_t$ to predict the resulting 3D scene flow. To evaluate performance, we derive the ground-truth trajectories from the current and subsequent images in the dataset. Our primary evaluation metric is the 3D Flow Error between the predicted and ground truth trajectories. While CoTracker samples points uniformly across the image, we observed that most pixels remain stationary. To ensure our metric reflects the important motion tracking accuracy and not be overly smoothed, we compute the error exclusively for pixels where the ground truth exhibits a displacement. Same as Eq. \ref{equ:flowalign}, we lift the track from CoTracker to 3D and compare with predicted 3D Flow. 

  The comparative results are summarized in Tab. \ref{tab:flow3d}. $\text{Act4R}_{\text{NSA}}$ deploys no spatial alignment in Sec. ~\ref{sec:fssa}, $\text{Act4R}_{\text{N3D}}$ does not use 3D flow loss, $\text{Act4R}_{\text{Gr00t}}$ uses an action encoder structure from Gr00t \citep{bjorck2025gr00t}, which only self-attends on action signal with no spatial context. 
  As shown in Tab. \ref{tab:flow3d}, we observe that Act4R significantly outperforms $\text{Act4R}_{\text{Gr00t}}$, demonstrating the effectiveness of fusing action and spatial embeddings in SAE. Spatial alignment reduces the 3D flow error on all seven platforms.  
  Finally, compared to NJF~\citep{Li2025} and as visualized in Fig.~\ref{fig:track} in terms of 2D tracks, Act4R demonstrates significant improvement in flow accuracy. Compared to a neural field representation, which has to voxelize the entire space, a point-map representation on the robot surface can be more compact; furthermore, Act4R leverages spatial prior from 4D VFMs whereas NJF optimizes neural fields from scratch, leading to improvement in performance while reducing the training time. We also provide evaluations for 2D pixel track error in the Appendix.

\subsection{Inverse Control Accuracy}
 \textbf{Inverse control based on 2D tracks.} 
 In the  dataset, each frame includes an input action attribute representing the robot's configuration. 
  We compute $\mathbf{A}_t$ as the difference in the action space between two consecutive frames. 
  To ensure numerical stability
  , we perform a linear normalization of these action vectors into a range of $[-1, 1]$. 
  This normalization prevents gradient imbalances caused by varying actuator scales across the seven different robotic platforms. Based on the optimization objective in Eq. \ref{eq:inverse_2D}, we extract 2D flow $\mathcal{S}_{tgt}$ by computing the track between $I_t$ and $I_{t+1}$ and use it to obtain optimal control $A^*_t$. 
  The experiment results are shown in Tab.\ref{tab:robot-performance}, where we use cosine similarity between  $A^*_t$ and groud truth action $\hat{A}^*_t$ to measure direction correctness. $\text{Act4R}_{\text{VGGT}}$ is the experiment that uses depth estimated by VGGT~\citep{Wang25vggt}. Magnitude is evaluated by  $\ell{2}$ error in the Appendix. With better scene flow estimation, Act4R outperform NJF~\citep{Li2025} on all platforms in terms of obtaining the correct actions to drive the robot.

 \textbf{Inverse control based on 3D tracks.} While 2D tracks are easy to obtain based on camera observations and consistent with prior evaluation protocol~\citep{Li2025}, robots move in 3D and should be able to follow 3D guidance. To this end, we also evaluate inverse control based on 3D tracks, which can be obtained by back-project 2D tracks into 3D space based on ground truth depth. We can then optimize Eq.~\ref{eq:inverse_3D} for the desired actions. Despite reasonable performance in following 2D tracks, $\text{Act4R}_{\text{N3D}}$ completely fails inverse control from 3D tracks, achieving only 0.085 on average in action accuracy, while Act4R and NJF~\cite{Li2025} maintain similar performance. $\text{Act4R}_{\text{VGGT}}$ performs worse than Act4R; this may be due to inconsistent depth estimation that leads to noise in optimization. A detailed table is included in the Appendix.

\begin{table}[!ht]
  \centering
  \small
  \begin{tabular}{l cccccc}
    \toprule
    & \multicolumn{6}{c}{Action Cosine Similarity $\uparrow$} \\
    \cmidrule{2-7}
    Robot & Act4R & $\text{Act4R}_{\text{VGGT}}$ & $\text{Act4R}_{\text{NSA}}$ & $\text{Act4R}_{\text{N3D}}$ & $\text{Act4R}_{\text{Gr00t}}$ & NJF~\citep{Li2025} \\
    \midrule
    Allegro        & \textbf{0.679} & 0.653 & 0.639 & 0.649 & 0.392 & 0.269 \\
    HSA            & 0.493 & 0.492 & 0.488 & 0.496 & \textbf{0.498} & 0.290 \\
    Toy Arm        & \textbf{0.536} & 0.531 & 0.526 & 0.529 & 0.501 & 0.273 \\
    Pneumatic      & \textbf{0.376} & 0.373 & 0.371 & 0.365 & 0.329 & 0.243 \\
    Allegro Sim    & \textbf{0.475} & 0.429 & 0.115 & 0.426 & 0.121 & 0.177 \\
    Franka Sim        & \textbf{0.399} & 0.384 & 0.355 & 0.383 & 0.288 & 0.185 \\
    UR5e Sim          & \textbf{0.623} & 0.616 & 0.603 & 0.616 & 0.447 & 0.503 \\
    \midrule
    Mean           & \textbf{0.512} & 0.497 & 0.442 & 0.495 & 0.368 & 0.277 \\
    \bottomrule
  \end{tabular}
  \caption{Performance comparison of action cosine similarity.}
  \label{tab:robot-performance}
\end{table}

\textbf{Closed-loop control.} Beyond evaluation of single-step inverse control, we want to understand if Act4R is generalizable and can successfully move robots across space through a sequence of actions. Real robots are often inaccessible and hard to be configured consistently across experiments; to this end, we evaluate long-sequence control through three simulated robotic embodiments: Allegro, Franka, and UR5e. We note that the kinematics of these robots are unknown to the system. 

\begin{figure}[htbp]
    \centering
    \setlength{\tabcolsep}{2pt}
    \renewcommand{\arraystretch}{0.6}
    \begin{tabular}{>{\raggedright\arraybackslash}p{1.35cm} cccc}
        & \small \textbf{GT} & \small \textbf{Act4R} & \small \textbf{$\text{Act4R}_{\text{VGGT}}$} & \small \textbf{NJF} \\[1pt]

        \rowlabel{Allegro} &
        \includegraphics[width=0.20\textwidth]{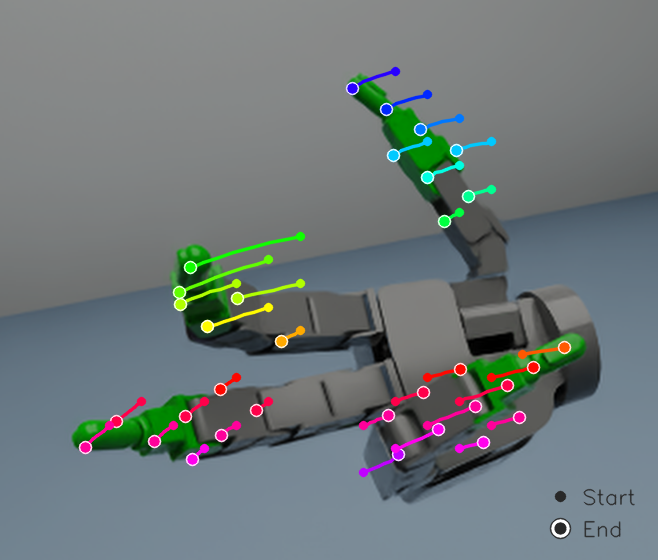} &
        \includegraphics[width=0.20\textwidth]{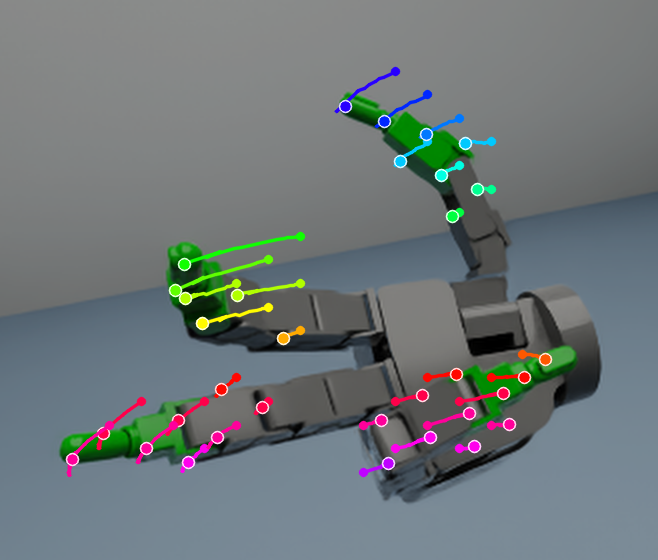} &
        \includegraphics[width=0.20\textwidth]{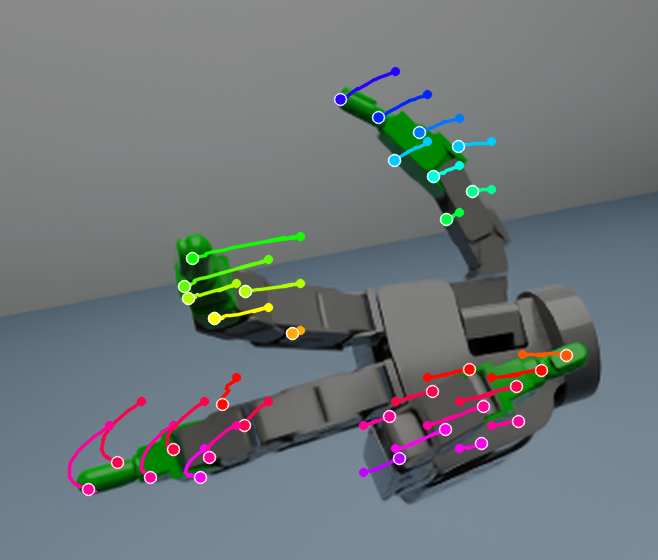} &
        \includegraphics[width=0.20\textwidth]{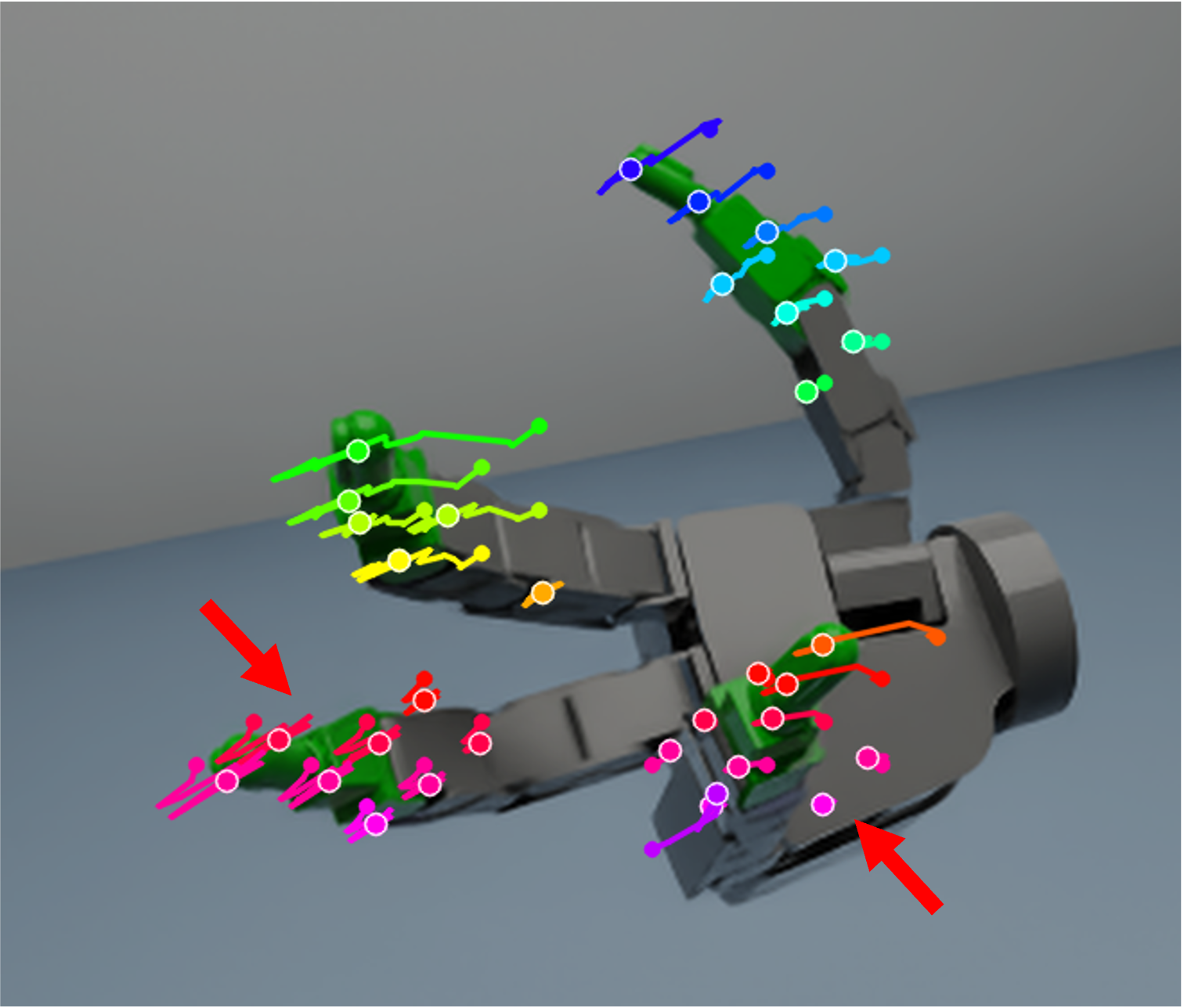} \\[-1pt]
        & \scriptsize  & \scriptsize \textbf{5.032} & \scriptsize 5.437 & \scriptsize 10.643 \\[4pt]

        \rowlabel{Franka} &
        \includegraphics[width=0.20\textwidth]{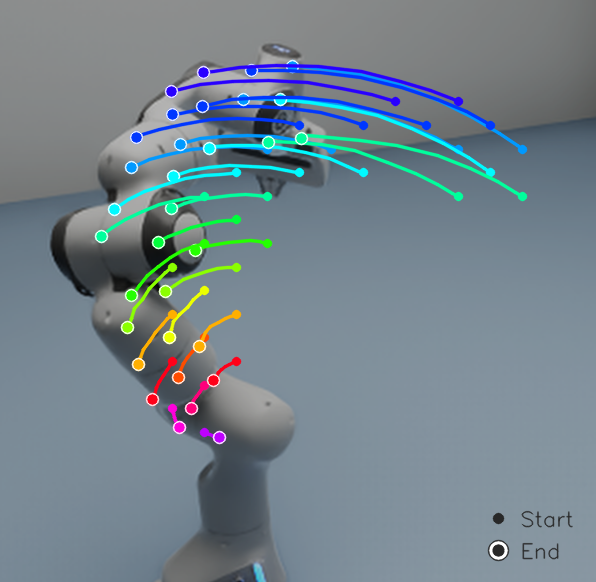} &
        \includegraphics[width=0.20\textwidth]{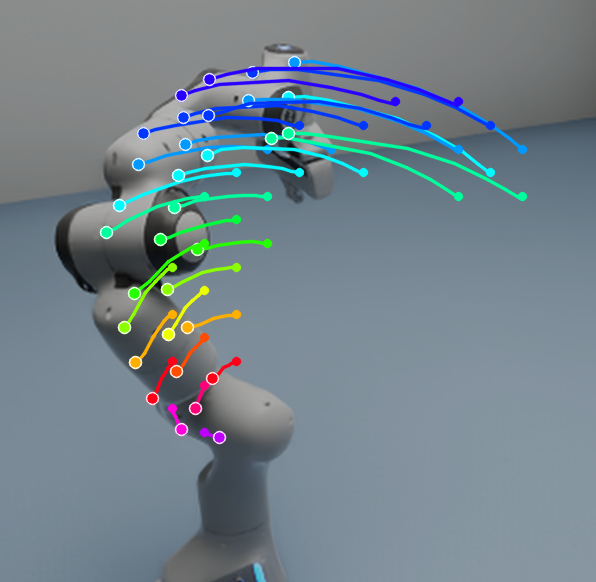} &
        \includegraphics[width=0.20\textwidth]{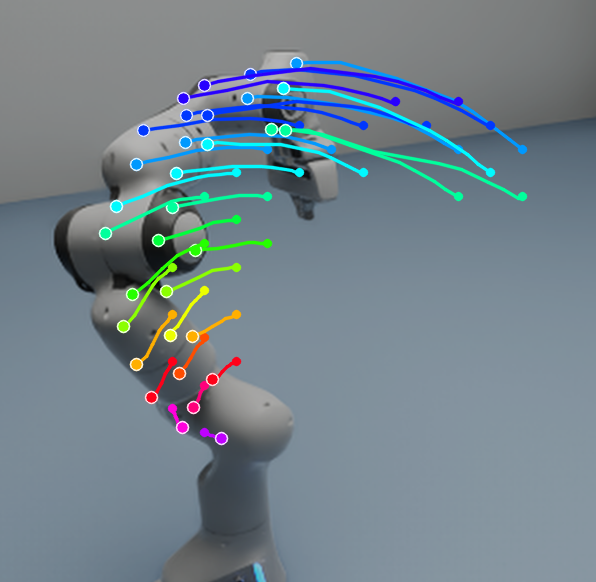} &
        \includegraphics[width=0.20\textwidth]{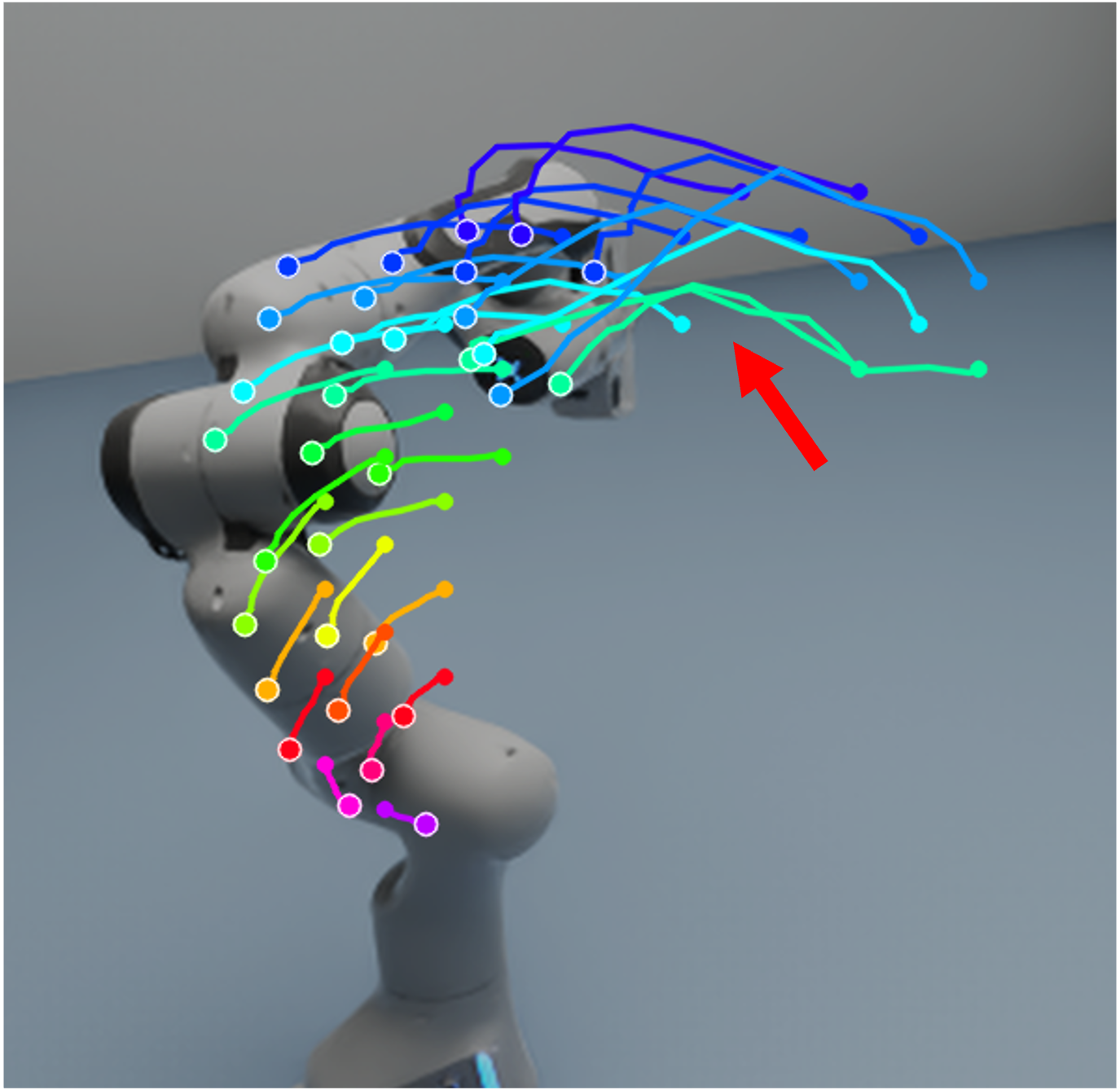} \\[-1pt]
        & \scriptsize  & \scriptsize \textbf{4.401} & \scriptsize 5.315 & \scriptsize 8.625 \\[4pt]

        \rowlabel{UR5e} &
        \includegraphics[width=0.20\textwidth]{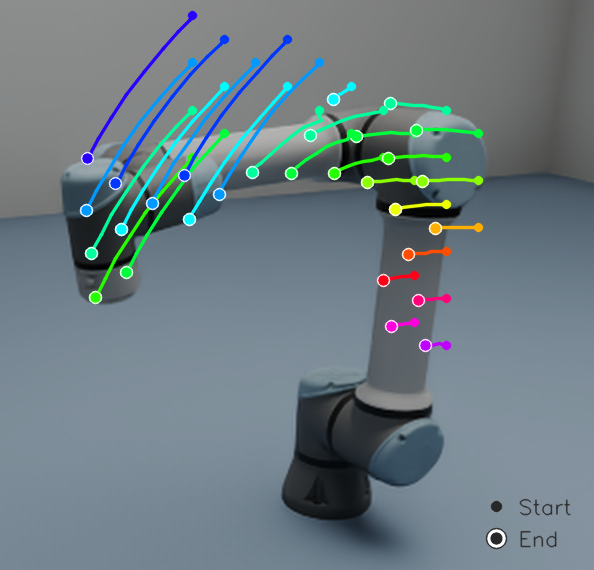} &
        \includegraphics[width=0.20\textwidth]{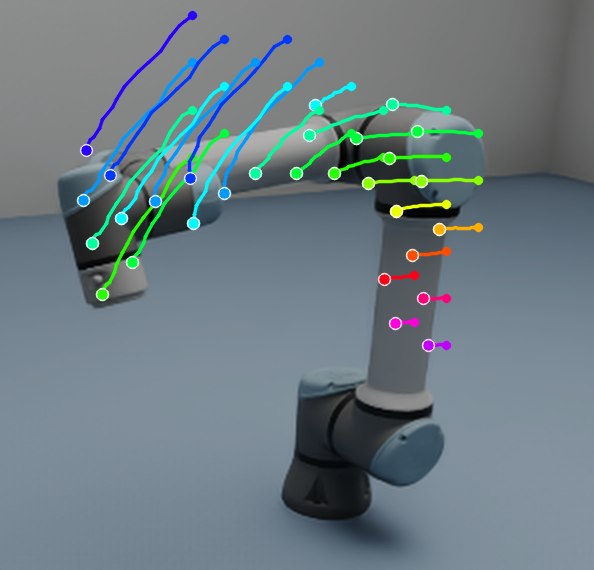} &
        \includegraphics[width=0.20\textwidth]{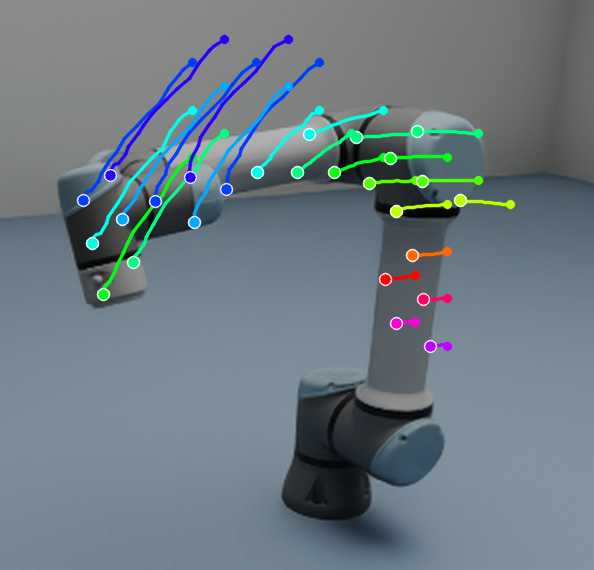} &
        \includegraphics[width=0.20\textwidth]{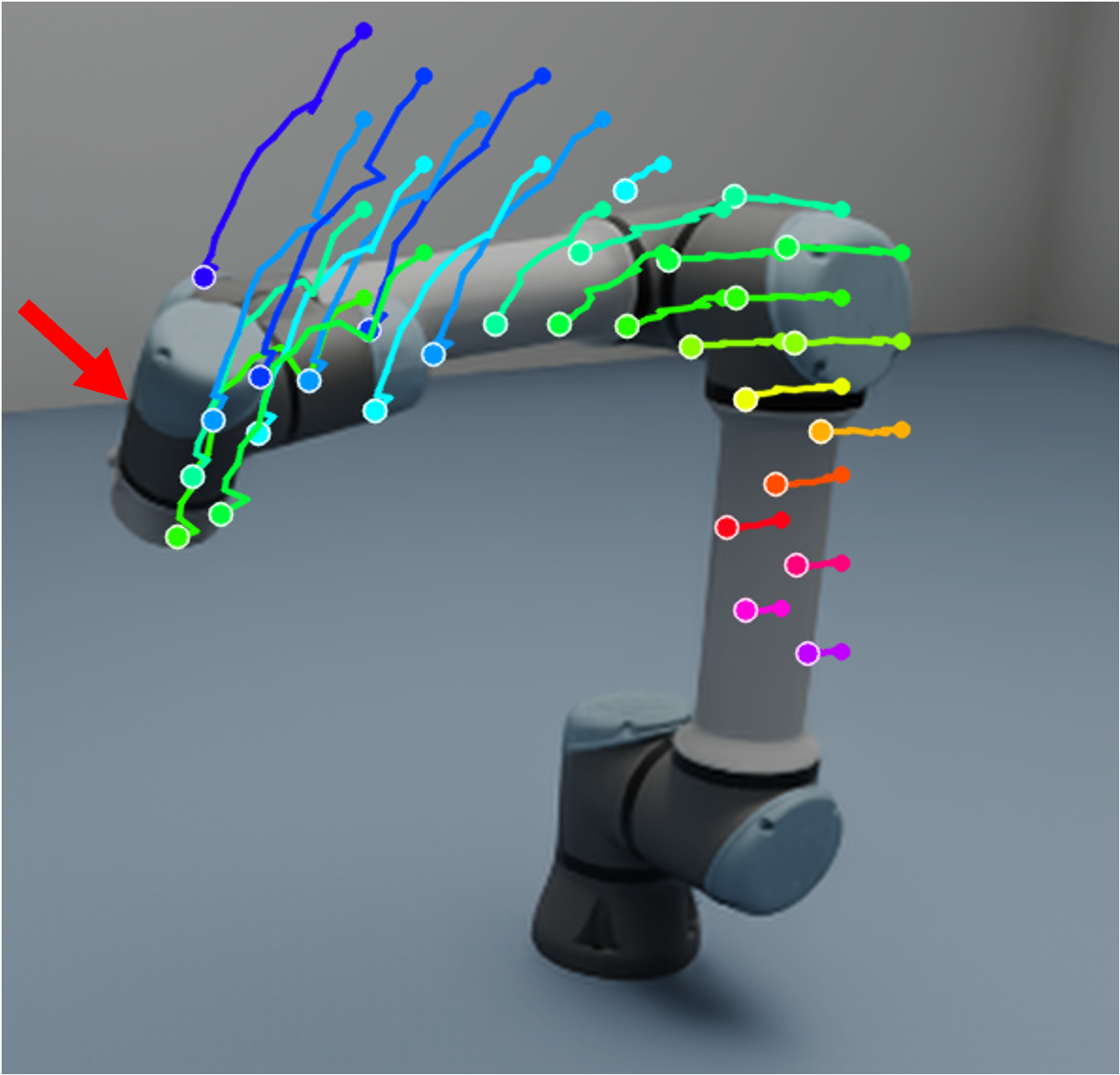} \\[-1pt]
        & \scriptsize CJPE & \scriptsize \textbf{6.380} & \scriptsize 6.928 & \scriptsize 8.885 \\[-2pt]
    \end{tabular}
    \caption{Long sequence inverse control visualization on simulated robots. GT is the ground-truth motion, Act4R is our method, $\text{Act4R}_{\text{VGGT}}$ uses VGGT depth, and NJF is the baseline. CJPE stands for Cumulative Joint Position Error, measured in radians.}
    \label{fig:simvis}
\end{figure}

To ensure that the desired control is achievable, we first generate a sequence of ground truth movement frames. We select the next ground-truth frame as target each step. Similar to before, we estimate the desired flow from the current observation of the robot with target frame and obtain the desired control through Eq.~\ref{eq:inverse_2D}. This process is repeated till robots observation is close enough with target, then advance to the next ground-truth frame.  As visualized in Fig.~\ref{fig:simvis}, Act4R can faithfully follow the instruction provided by GT tracks, while NJF~\citep{Li2025} often leads to incomplete movement (the pinky finger in Allegro) or very noisy movement (Franka and UR5e). 
We measure the trajectory accuracy by the cumulative joint position error:
\begin{equation}
     E_{\mathrm{CJPE}} = \sum_{t=1}^{T}\sum_{i=1}^{J}\left| q_{i,t} - q^{\mathrm{gt}}_{i,t} \right|
\end{equation}

where $q^{\mathrm{gt}}_{i,t}$ is the ground-truth position of joint $i$ at frame $t$ in radians, and $q_{i,t}$ is the corresponding joint position once the observation is close enough to that target, before advancing to the next frame. $T$ is the number of ground-truth frames and $J$ is the number of joints.
Interestingly, Act4R trained on VGGT-estimated depth instead of ground truth depth can still lead to similar performance, despite slightly lower action accuracies in Tab.~\ref{tab:robot-performance}. For more visualizations of closed-loop control, we refer to Fig.~\ref{fig:intro_examples}, where various robots can be successfully guided to draw desired letters. 

\section{Conclusion and Discussion}
\label{sec:conclusion}
\vspace{-5pt}
  We propose Act4R, an algorithm designed to estimate future movements of robots given current 2D observation and an action signal. Act4R leverages strong spatial prior from 3D VFM by inheriting its encoder-decoder design; particularly, Act4R introduces a Spatial-Action Encoder to capture the spatial-action relationship. To effectively learn the future spatial embedding, SAE models the Value Token from the joint embedding of action and spatial embeddings; furthermore, we propose a Future State Spatial Alignment strategy to help SAE learn more effectively. Once trained, our model can be used to infer the required action given a desired scene flow through optimization. Experiments show that our method significantly outperforms NJF~\citep{Li2025} in scene flow prediction while incurring lower training cost. 
  If measured depth is unavailable, depth predicted by a vision model can still be used.
  Despite its promising performance, Act4R still has room to improve.  When the spatial displacement between images is small, vision-based inverse control can be noisy. Furthermore, more modern 4D Foundation Models~\citep{luo20264rc,zhang2026efficiently} are available as the backbone, though at much higher cost. Finally, the Act4R formulation can be used to model more complex interactions, e.g., with soft tissues. Future work will focus on addressing these challenges and evaluate on more complex settings.
  


\newpage
\subsection*{AI use statement}
In this work, we used generative AI tools to improve the language, clarity, and readability of the manuscript. All scientific contributions, including the core ideas, methodology, experimental design, and interpretation of the results, are entirely the original work of the authors. All authors have carefully reviewed the final manuscript and take full responsibility for its content and claims.

\bibliography{iclr2027_conference}
\bibliographystyle{iclr2027_conference}

\appendix
\newpage
\section{Appendix}
\subsection{Additional Loss}

  \textbf{Depth Loss.} 
  To supervise 3D geometry, we fetch the depth $z$ dimension of point cloud $\mathbf{X}_{t+1}$ and use reference depth $z_{gt}$:
  \begin{equation}
      \mathcal{L}_{depth} = \frac{1}{N} \sum_{i=1}^{N} \left\|\alpha \cdot z_{i} - z_{gt,i}\right\|^2, \alpha = \frac{\sum (z \cdot z_{gt})}{\sum z^2}.
  \end{equation}


  \textbf{Consistency Loss.} 
  To ensure temporal stability and maintain geometric consistency across the predicted states, we introduce a consistency loss $\mathcal{L}_{cons}$.
  Let $\mathbf{X'}_{t}$ and $\mathbf{X'}_{t+1}$ denote the predicted 3D coordinates of corresponding points at static points of both time steps. 
  The consistency objective is defined as:
  \begin{equation}
      \mathcal{L}_{cons} = \sum_{n=1}^{N} \left\| \mathbf{X'}_{t} - \mathbf{X'}_{t+1} \right\|_2
  \end{equation}
  This loss ensures that the geometry remains invariant across the predicted transition, forcing the model to represent the robot's motion as a displacement of a single, coherent 3D structure rather than two independent reconstructions.

\subsection{2D Pixel flow}
Tab.~\ref{tab:mpe} reports the 2D pixel error. All methods are evaluated on the same tracked pixels, and the error is the distance between the predicted and ground-truth endpoints, measured in pixels. 

Under the 2D track metric, Act4R achieves the lowest error, and is the best method on five of seven robots. On HSA it is only slightly worse, which may come from the small displacement in that dataset. On Allegro Sim, $\text{Act4R}_{\text{NSA}}$ is better, and Act4R is nearly tied with NJF. The gap to NJF is large on the soft robots.
Adding the 3D flow loss lowers the mean from 1.579 to 1.485, which indicates the 3D loss also improves 2D performance.
\begin{table}[!ht]
  \centering
  \small
  \begin{tabular}{l ccccc}
    \toprule
    & \multicolumn{5}{c}{Track Mean Pixel Error $\downarrow$} \\
    \cmidrule{2-6}
    Robot & Act4R & $\text{Act4R}_{\text{NSA}}$ & $\text{Act4R}_{\text{N3D}}$ & $\text{Act4R}_{\text{Gr00t}}$ & NJF~\citep{Li2025} \\
    \midrule
    Allegro        & \textbf{1.959} & 2.173 & 2.127 & 5.309 & 3.618 \\
    HSA            & 1.712 & 1.727 & \textbf{1.710} & \textbf{1.710} & 4.768 \\
    Toy Arm        & \textbf{1.433} & 1.490 & 1.492 & 1.629 & 2.683 \\
    Pneumatic      & \textbf{1.502} & 1.534 & 1.597 & 1.665 & 4.105 \\
    Allegro Sim    & 1.620 & \textbf{1.487} & 1.683 & 3.478 & 1.628 \\
    Franka Sim        & \textbf{1.142} & 1.890 & 1.354 & 2.504 & 1.950 \\
    UR5e Sim          & \textbf{1.028} & 1.098 & 1.093 & 2.260 & 1.388 \\
    \midrule
    Mean           & \textbf{1.485} & 1.628 & 1.579 & 2.651 & 2.877 \\
    \bottomrule
  \end{tabular}
  \caption{Performance comparison of track mean pixel error.}
  \label{tab:mpe}
\end{table}

\begin{figure}[tbp]
    \centering
    \setlength{\tabcolsep}{3pt}
    \renewcommand{\arraystretch}{0}

    \begin{tabular}{m{0.6cm} cccccc}
        & \small \textbf{GT} & \small \textbf{$\text{Act4R}_{\text{VGGT}}$} & \small \textbf{$\text{Act4R}_{\text{NSA}}$} & \small \textbf{$\text{Act4R}_{\text{N3D}}$} & \small \textbf{$\text{Act4R}_{\text{Gr00t}}$} & \small \textbf{Act4R (Ours)} \\[1.5ex]

        \rotatebox[origin=c]{90}{\small \textbf{Allegro}} &
        \begin{subfigure}[c]{0.145\textwidth}
            \centering
            \includegraphics[width=\linewidth]{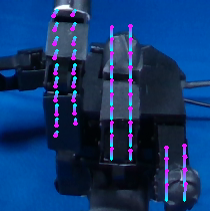}\\[-3pt]
            \scriptsize \phantom{0.000}
        \end{subfigure} &
        \begin{subfigure}[c]{0.145\textwidth}
            \centering
            \includegraphics[width=\linewidth]{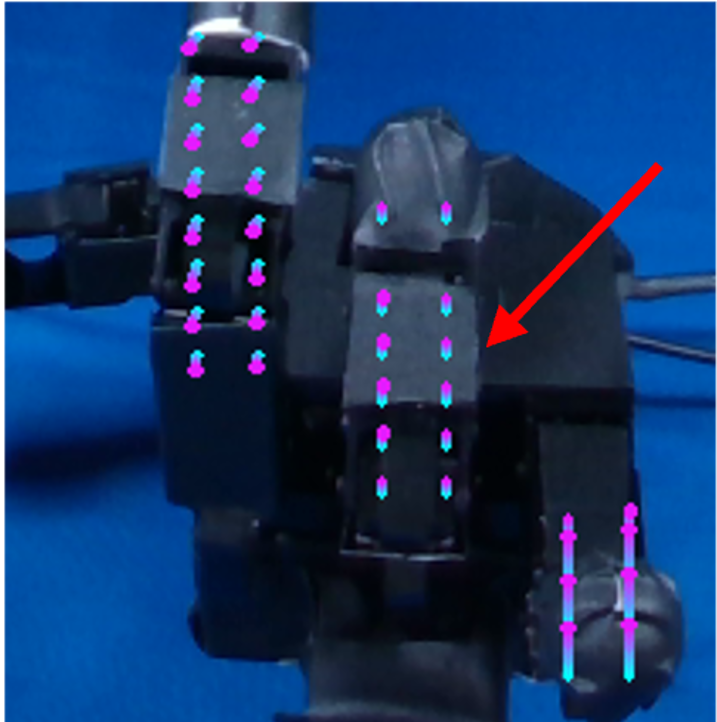}\\[-3pt]
            \scriptsize 3.158
        \end{subfigure} &
        \begin{subfigure}[c]{0.145\textwidth}
            \centering
            \includegraphics[width=\linewidth]{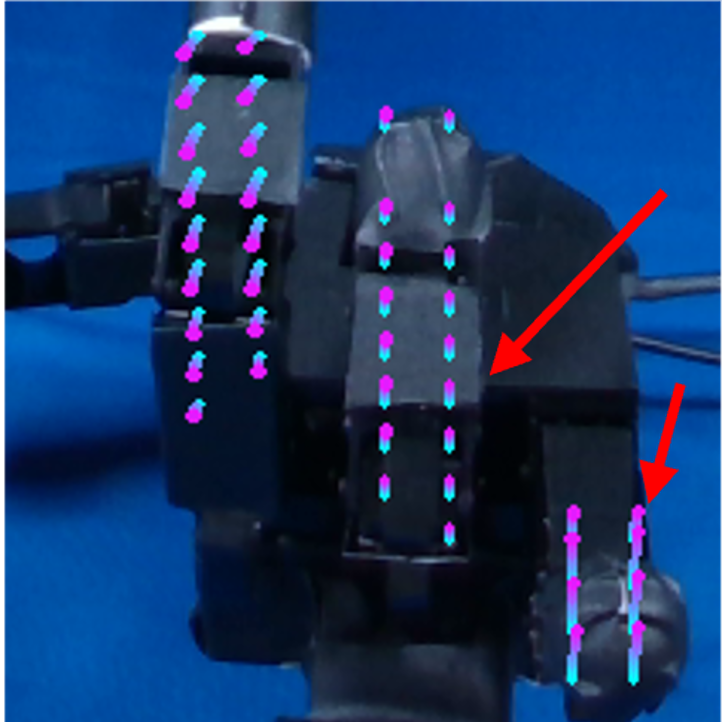}\\[-3pt]
            \scriptsize 2.134
        \end{subfigure} &
        \begin{subfigure}[c]{0.145\textwidth}
            \centering
            \includegraphics[width=\linewidth]{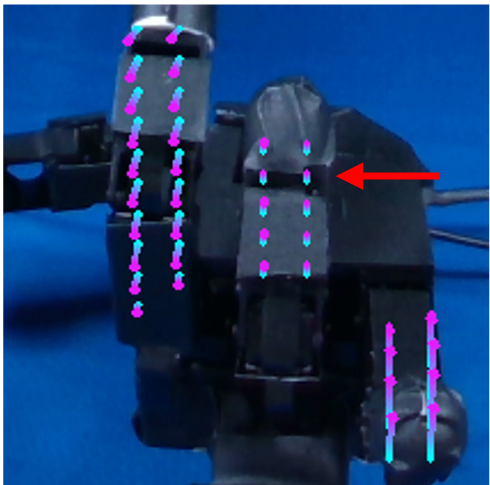}\\[-3pt]
            \scriptsize 3.100
        \end{subfigure} &
        \begin{subfigure}[c]{0.145\textwidth}
            \centering
            \includegraphics[width=\linewidth]{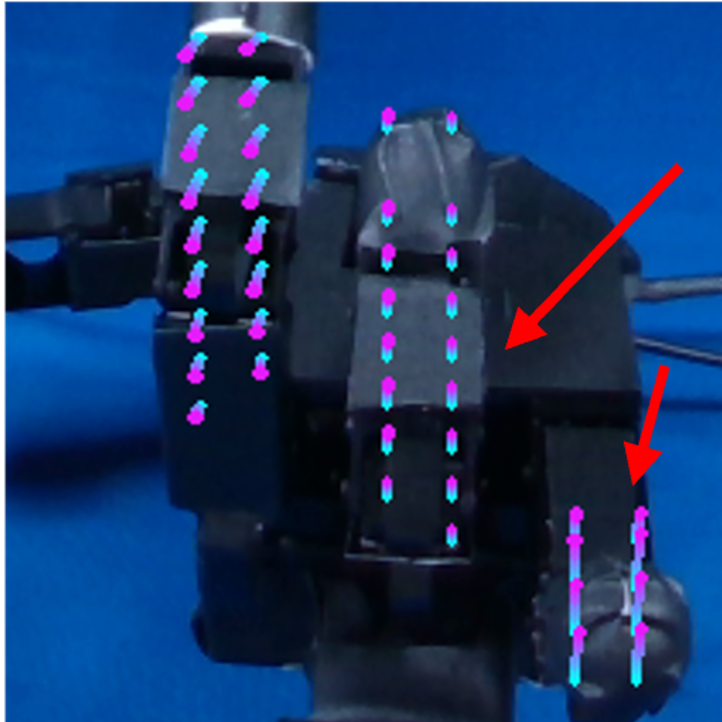}\\[-3pt]
            \scriptsize 4.106
        \end{subfigure} &
        \begin{subfigure}[c]{0.145\textwidth}
            \centering
            \includegraphics[width=\linewidth]{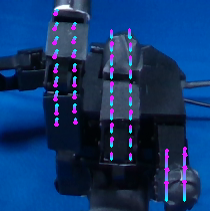}\\[-3pt]
            \scriptsize \textbf{2.132}
        \end{subfigure} \\[2.5ex]

        \rotatebox[origin=c]{90}{\small \textbf{HSA}} &
        \begin{subfigure}[c]{0.145\textwidth}
            \centering
            \includegraphics[width=\linewidth]{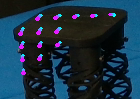}\\[-3pt]
            \scriptsize \phantom{0.000}
        \end{subfigure} &
        \begin{subfigure}[c]{0.145\textwidth}
            \centering
            \includegraphics[width=\linewidth]{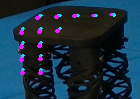}\\[-3pt]
            \scriptsize \textbf{1.007}
        \end{subfigure} &
        \begin{subfigure}[c]{0.145\textwidth}
            \centering
            \includegraphics[width=\linewidth]{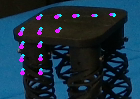}\\[-3pt]
            \scriptsize 1.057
        \end{subfigure} &
        \begin{subfigure}[c]{0.145\textwidth}
            \centering
            \includegraphics[width=\linewidth]{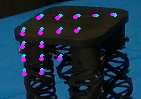}\\[-3pt]
            \scriptsize 1.042
        \end{subfigure} &
        \begin{subfigure}[c]{0.145\textwidth}
            \centering
            \includegraphics[width=\linewidth]{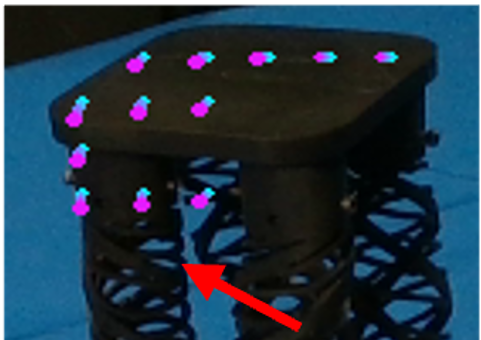}\\[-3pt]
            \scriptsize 1.257
        \end{subfigure} &
        \begin{subfigure}[c]{0.145\textwidth}
            \centering
            \includegraphics[width=\linewidth]{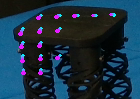}\\[-3pt]
            \scriptsize 1.032
        \end{subfigure} \\[2.5ex]

        \rotatebox[origin=c]{90}{\small \textbf{ToyArm}} &
        \begin{subfigure}[c]{0.145\textwidth}
            \centering
            \includegraphics[width=\linewidth]{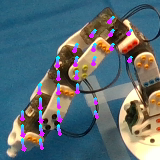}\\[-3pt]
            \scriptsize \phantom{0.000}
        \end{subfigure} &
        \begin{subfigure}[c]{0.145\textwidth}
            \centering
            \includegraphics[width=\linewidth]{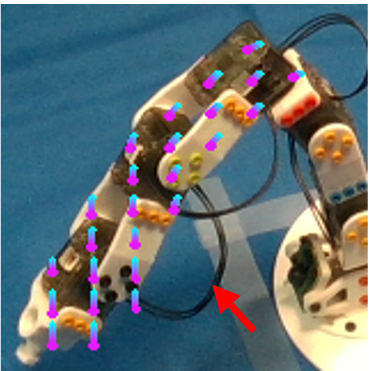}\\[-3pt]
            \scriptsize 1.625
        \end{subfigure} &
        \begin{subfigure}[c]{0.145\textwidth}
            \centering
            \includegraphics[width=\linewidth]{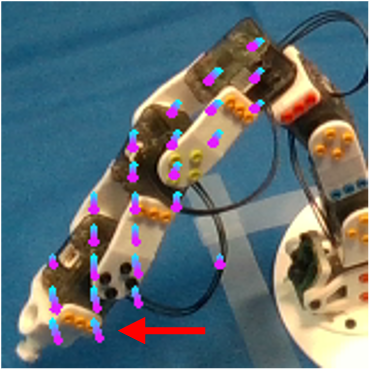}\\[-3pt]
            \scriptsize 1.715
        \end{subfigure} &
        \begin{subfigure}[c]{0.145\textwidth}
            \centering
            \includegraphics[width=\linewidth]{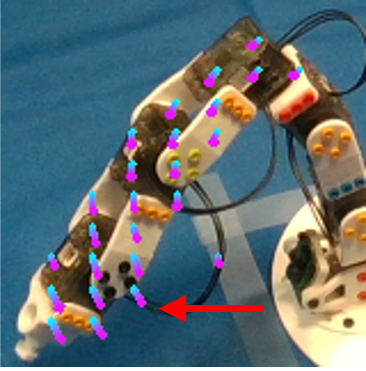}\\[-3pt]
            \scriptsize 2.181
        \end{subfigure} &
        \begin{subfigure}[c]{0.145\textwidth}
            \centering
            \includegraphics[width=\linewidth]{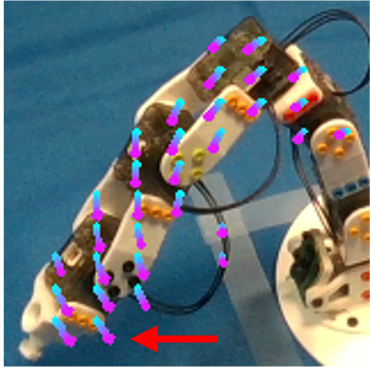}\\[-3pt]
            \scriptsize 1.747
        \end{subfigure} &
        \begin{subfigure}[c]{0.145\textwidth}
            \centering
            \includegraphics[width=\linewidth]{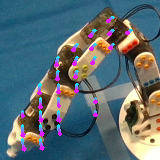}\\[-3pt]
            \scriptsize \textbf{1.614}
        \end{subfigure} \\[2.5ex]

        \rotatebox[origin=c]{90}{\small \textbf{Pneumatic}} &
        \begin{subfigure}[c]{0.145\textwidth}
            \centering
            \includegraphics[width=\linewidth]{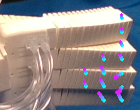}\\[-3pt]
            \scriptsize \phantom{0.000}
        \end{subfigure} &
        \begin{subfigure}[c]{0.145\textwidth}
            \centering
            \includegraphics[width=\linewidth]{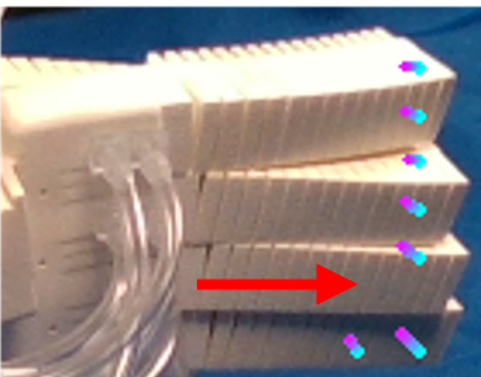}\\[-3pt]
            \scriptsize 5.113
        \end{subfigure} &
        \begin{subfigure}[c]{0.145\textwidth}
            \centering
            \includegraphics[width=\linewidth]{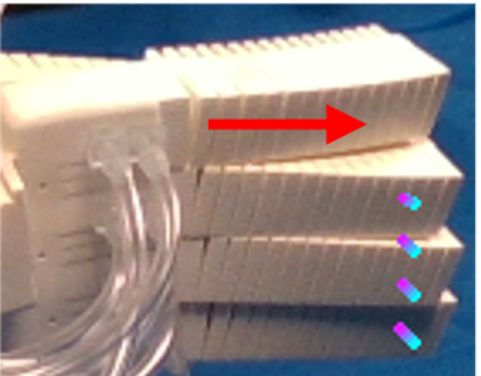}\\[-3pt]
            \scriptsize 5.422
        \end{subfigure} &
        \begin{subfigure}[c]{0.145\textwidth}
            \centering
            \includegraphics[width=\linewidth]{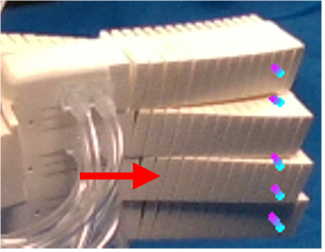}\\[-3pt]
            \scriptsize 5.655
        \end{subfigure} &
        \begin{subfigure}[c]{0.145\textwidth}
            \centering
            \includegraphics[width=\linewidth]{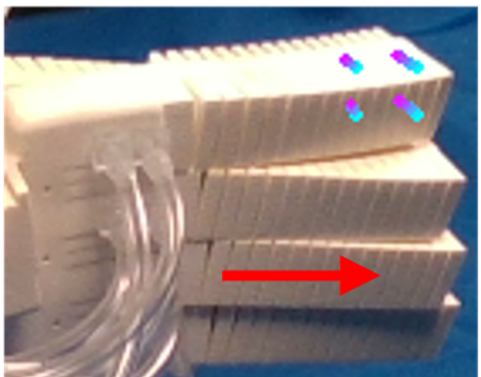}\\[-3pt]
            \scriptsize 6.053
        \end{subfigure} &
        \begin{subfigure}[c]{0.145\textwidth}
            \centering
            \includegraphics[width=\linewidth]{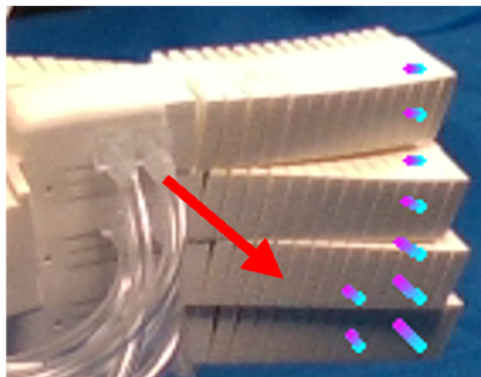}\\[-3pt]
            \scriptsize \textbf{4.952}
        \end{subfigure} \\
    \end{tabular}
    \caption{Visualization results for the ablation study. Values below each image indicate the Mean Pixel Error.}
    \label{fig:ablation}
\end{figure}

Fig.~\ref{fig:ablation} shows the track prediction under different experimental settings.

\subsection{Additional Action Metrics}

Cosine similarity does not capture magnitude error. Tab. \ref{tab:action-l2} shows the $\ell{2}$ error metric. On HSA, $\text{Act4R}{\text{Gr00t}}$ is slightly lower, and on the pneumatic hand, $\text{Act4R}{\text{NSA}}$ is a little bit better compared with Act4R. NJF is much worse on every robot. Replacing measured depth with VGGT depth raises the mean only from 0.121 to 0.123. This suggests estimated depth is a viable substitute if measured depth is unavailable.

We also recover actions from 3D scene flow using Eq.~\ref{eq:inverse_3D} and report the cosine similarity in Tab.~\ref{tab:action-cossim-3d}. Act4R achieves the highest average similarity and performs best on six of the seven robots, while NJF slightly outperforms it on HSA. Overall similarities are lower than those obtained from 2D tracks (Tab.~\ref{tab:robot-performance}), with the largest degradation observed on Allegro Sim. Replacing ground-truth depth with VGGT depth reduces the mean similarity from 0.446 to 0.377. Furthermore, removing the 3D flow loss causes $\text{Act4R}_\text{N3D}$ to drop sharply to 0.085, highlighting the importance of this loss when the target motion is specified in 3D.

\begin{table}[!ht]
  \centering
  \small
  \begin{tabular}{l cccccc}
    \toprule
    & \multicolumn{6}{c}{Action $\ell{2}$ Error $\downarrow$} \\
    \cmidrule{2-7}
    Robot & Act4R & $\text{Act4R}_{\text{VGGT}}$ & $\text{Act4R}_{\text{NSA}}$ & $\text{Act4R}_{\text{N3D}}$ & $\text{Act4R}_{\text{Gr00t}}$ & NJF~\citep{Li2025} \\
    \midrule
    Allegro        & \textbf{0.129} & 0.134 & 0.136 & 0.135 & 0.167 & 0.900 \\
    HSA            & 0.158 & 0.159 & 0.162 & 0.158 & \textbf{0.155} & 1.231 \\
    Toy Arm        & \textbf{0.162} & \textbf{0.162} & 0.163 & 0.163 & 0.166 & 0.965 \\
    Pneumatic      & 0.229 & 0.229 & \textbf{0.228} & 0.231 & 0.234 & 1.220 \\
    Allegro Sim    & \textbf{0.069} & 0.070 & 0.278 & 0.073 & 0.261 & 0.644 \\
    Franka Sim        & \textbf{0.064} & 0.066 & 0.068 & 0.067 & 0.075 & 0.485 \\
    UR5e Sim          & \textbf{0.037} & 0.040 & 0.051 & 0.045 & 0.041 & 0.093 \\
    \midrule
    Mean           & \textbf{0.121} & 0.123 & 0.155 & 0.125 & 0.157 & 0.791 \\
    \bottomrule
  \end{tabular}
  \caption{Performance comparison of action $\ell{2}$ error.}
  \label{tab:action-l2}
\end{table}

\begin{table}[!ht]
  \centering
  \small
  \begin{tabular}{l cccc}
    \toprule
    & \multicolumn{4}{c}{Action Cosine Similarity 3D $\uparrow$} \\
    \cmidrule{2-5}
    Robot & Act4R & $\text{Act4R}_{\text{VGGT}}$ & $\text{Act4R}_{\text{N3D}}$ & NJF~\citep{Li2025} \\
    \midrule
    Allegro        & \textbf{0.665} & 0.573 & 0.104 & 0.523 \\
    HSA            & 0.489 & 0.427 & 0.040 & \textbf{0.503} \\
    Toy Arm        & \textbf{0.570} & 0.470 & 0.230 & 0.354 \\
    Pneumatic      & \textbf{0.302} & 0.268 & 0.067 & 0.130 \\
    Allegro Sim    & \textbf{0.149} & 0.115 & 0.060 & 0.072 \\
    Franka Sim        & \textbf{0.396} & 0.301 & 0.036 & 0.063 \\
    UR5e Sim          & \textbf{0.551} & 0.483 & 0.061 & 0.191 \\
    \midrule
    Mean           & \textbf{0.446} & 0.377 & 0.085 & 0.262 \\
    \bottomrule
  \end{tabular}
  \caption{Performance comparison of 3D action cosine similarity.}
  \label{tab:action-cossim-3d}
\end{table}

\end{document}